\documentclass[11pt,a4paper]{article}
\usepackage[hyperref]{acl2021}
\usepackage{times}
\usepackage{latexsym}
\usepackage{amsmath}
\usepackage{amsfonts}
\usepackage{booktabs}
\usepackage{tikz}
\usepackage{subfigure}
\usepackage{caption}

\usepackage{microtype}
\aclfinalcopy

\newcommand{\pixeltocap}{\textbf{Pixel2Caption}}
\newcommand{\pixeltocapatt}{\textbf{Pixel2Caption+\textit{att}}}
\newcommand{\glstm}{\textbf{G-LSTM}}
\newcommand{\glstmatt}{\textbf{G-LSTM+\textit{att}}}
\newcommand{\glstmenc}{\textbf{G-LSTM+\textit{enc}}}
\newcommand{\glstmencatt}{\textbf{G-LSTM+\textit{enc}+\textit{att}}}

\newcommand\blfootnote[1]{%
  \begingroup
  \renewcommand\thefootnote{}\footnote{#1}
  \addtocounter{footnote}{-1}
  \endgroup
}

\author{
  \textbf{Maximilian Mozes}$^{1,3}$ \quad \textbf{Martin Schmitt}$^{2}$ \quad \textbf{Vladimir Golkov}$^{1}$ \quad \\
  \textbf{Hinrich Sch\"utze}$^{2}$ \quad \textbf{Daniel Cremers}$^{1}$ \\
  $^{1}$Computer Vision Group, Technical University of Munich \\
  $^{2}$Center for Information and Language Processing (CIS), LMU Munich \\
  $^{3}$University College London \\ 
  \small{\texttt{maximilian.mozes@ucl.ac.uk}}
}

\title{Scene Graph Generation for Better Image Captioning?}

\begin{document}
\maketitle

\blfootnote{Technical report. Work done and written in 2019.}

\begin{abstract}
We investigate the incorporation of visual relationships into the task of supervised image caption generation by proposing a model that leverages detected objects and auto-generated visual relationships to describe images in natural language. To do so, we first generate a scene graph from raw image pixels by identifying individual objects and visual relationships between them. This scene graph then serves as input to our graph-to-text model, which generates the final caption. In contrast to previous approaches, our model thus explicitly models the detection of objects and visual relationships in the image. For our experiments we construct a new dataset from the intersection of Visual Genome and MS COCO, consisting of images with both a corresponding gold scene graph and human-authored caption. Our results show that our methods outperform existing state-of-the-art end-to-end models that generate image descriptions directly from raw input pixels when compared in terms of the BLEU and METEOR evaluation metrics. 
\end{abstract}

\section{Introduction}
Recent works dealing with the generation of text from data structures such as images~\cite[e.g.,][]{karpathy15,vinyals15}, videos~\cite[e.g.,][]{venugopalan15iccv} or audio~\cite[e.g.,][]{graves2013} have shown that supervised learning algorithms are capable of aligning semantic concepts across different modalities. In this work, we focus on the task of automatic image captioning, a widely-studied task at the intersection of vision and language research. Most approaches to image captioning operate by conditioning a decoder model on an abstracted representation of the input image instead of explicitly taking detected objects and visual relationships into account~\cite[e.g.,][]{karpathy15,xu16}. However, natural language descriptions in general and captions in particular are dominated by discrete objects standing in discrete relations. By forcing the generation process to go through a scene graph consisting of objects and relations, we impose an appropriate structural bias that is lacking in direct pixel-to-caption generation. We therefore approach the task of supervised image caption generation by developing an architecture that makes explicit use of detected visual objects and their semantic relationships in a given input image to generate an image description in natural language.  More specifically, our method consists of a two-step approach that first extracts a scene graph (i.e., objects and their visual relationships) from an input image and then utilizes this representation to generate an image description in natural language. In doing so, we incorporate an existing method for supervised scene graph generation, i.e., \textsc{MotifNet}~\cite{zellers2018scenegraphs}, to extract visual semantic concepts from images and represent them in form of scene graphs.

Scene graphs have been utilized in a variety of tasks such as image retrieval~\cite[e.g.,][]{johnson15} and image generation~\cite{johnson18} and are of particular interest for tasks dealing with the alignment of visual and textual concepts, since the representations utilize words to describe phenomena that are present in visual scenarios. While numerous approaches for image-to-graph generation and visual relationship detection have been proposed in recent years~\cite[e.g.,][]{lu16,newell17,li17,yang18,zellers2018scenegraphs,zhang2018large}, little attention has thus far been paid to the problem of graph-to-text generation. We hence propose a variety of methods utilizing recurrent neural network mechanisms operating on scene graphs for the generation of natural language and show that the presence of visual objects and their relationships is beneficial for the automatic description of images. 

Our work thus presents the following main contributions:
\begin{enumerate}
    \item We propose a two-step supervised learning approach that generates scene graphs from raw input pixels and utilizes these graph representations to generate image descriptions in natural language.
    \item We show that such a simple two-step approach outperforms conventional CNN-LSTM image captioning architectures.
\end{enumerate}

\section{Related work}
\label{sec:related_work}
The problem of end-to-end image caption generation has been studied widely in the context of deep learning in recent years. Pioneering approaches to this problem utilize a combination of convolutional and recurrent neural networks processing the visual and textual data representations, respectively. Multiple encoder-decoder approaches have been proposed that employ a CNN transforming a raw input image to a dense vector representation which is then used to condition a neural language model generating a descriptive sequence in natural language~\cite[e.g.,][]{chen15, donahue15,vinyals15,karpathy15,wang2016}. 

Building upon this idea,~\citet{xu16} propose the first approach to incorporate an additional attention mechanism into the model's decoder, enabling it to refer back to the abstracted image representation at each time step during the generation of an image caption. Subsequent approaches extend the incorporation of attention mechanisms for image captioning~\cite[e.g.,][]{yang2016,lu2017,khademi18}. For instance,~\citet{lu2017} extend the idea of incorporating visual attention to the image caption generation task by introducing an adaptive attention mechanism allowing the model to decide to what extent it should rely on the visual and linguistic features when generating an image caption.

\begin{figure*}[t]
\centering
\includegraphics[width=0.95\textwidth]{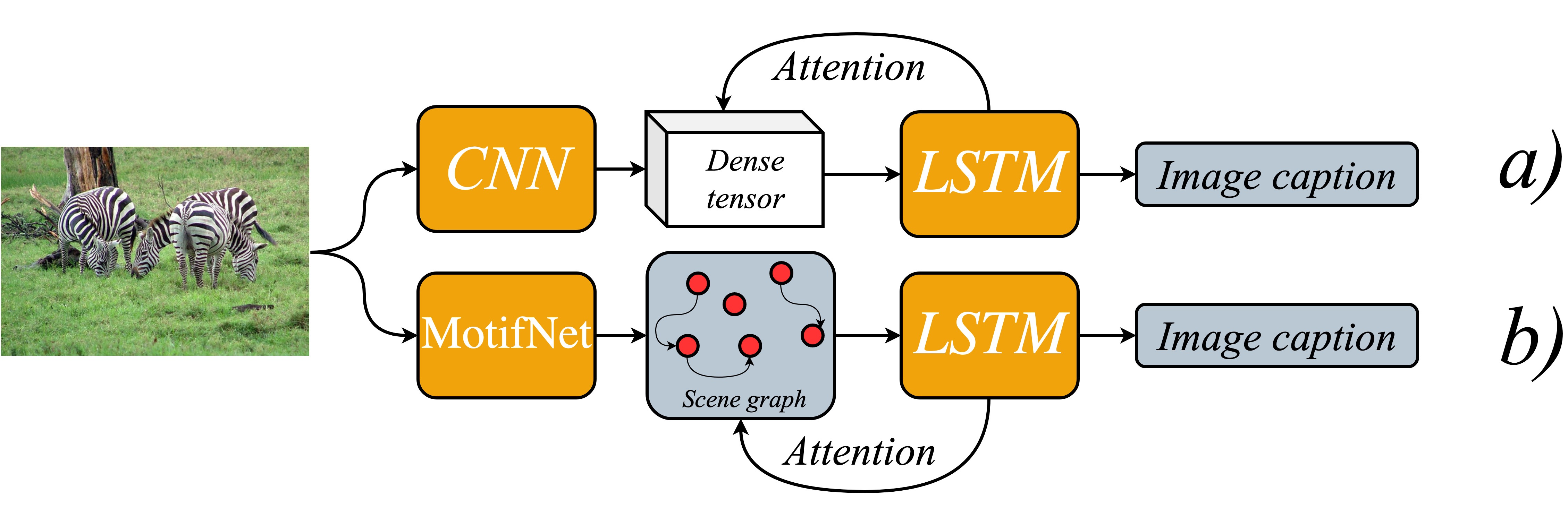}
\caption{Illustration and comparison of the \pixeltocapatt{} (described in \textit{\textbf{a)}}) and our \glstmatt{} (shown in \textit{\textbf{b)}}) methods. \pixeltocapatt{} processes the input image by computing a dense numerical representation using a CNN which is then used to condition an LSTM that generates the image caption. \glstmatt{}, in contrast, utilizes \textsc{MotifNet} to craft a scene graph from the input image, which is then processed by an LSTM to generate the final caption.}
\label{fig:architecture}
\end{figure*}

\subsection{Generating image captions from visual relationships}
Although comparatively little attention has been paid to the generation of image captions via visual relationships, there exists a variety of works employing these characteristics to generate image captions.

\citet{Yao2018ExploringVR} propose an architecture that utilizes region-based visual relationships to generate an image caption for a given image. Specifically, their method uses the \textit{Faster R-CNN}~\cite{ren2015} object detector to identify a set of objects present in an input image. Afterwards, a classification method is applied on pairs of detected objects to identify their most probable semantic relationship. The resulting graph representation is then forwarded to two Graph Convolutional Neural Networks (CGN) that generate relation-aware region features for all the detected regions based on their predicted visual relationships. Finally, a two-layer LSTM is conditioned on the region-level features generated by the CGN module, and generates the image caption based on this representation.~\citet{Yao2018ExploringVR} additionally install an attention mechanism in the LSTM decoder that operates over the region features at each time step when generating the output predictions. 

Moreover,~\citet{Yang2018AutoEncodingSG} propose the incorporation of scene graphs into image captioning by utilizing them to model language inductive bias during the task, and~\citet{kim2019dense} propose a dense captioning mechanism that produces multiple individual captions per image. Their approach initially uses a bounding box object detector that identifies object regions present in an input image. Afterwards, a recurrent neural network is trained to generate a caption for each relational pair of identified objects.

Two recent works published by~\citet{li2019know} and~\citet{hou2019relational} present approaches that are similar to our work.~\citet{li2019know} combine scene graphs for image captioning in conjunction with a hierarchical attention network. Their approach first uses a Region Proposal Network~\cite{girshick2015fast} to compute object proposals for an input image. These proposals are then used to generate both a visual feature representation and semantic relationship features, which are forwarded to an LSTM decoder with a hierarchical attention module generating the image caption.~\citet{hou2019relational} provide a different method for incorporating scene graphs into the image captioning pipeline by utilizing scene graphs sourced from the Visual Genome dataset as external prior knowledge graphs.

\section{Method}
The proposed approach for generating image captions via visual relationships is divided into two parts. Our model tackles the \textit{image-to-text} generation task by first generating an intermediate scene graph representation of the input image and then decodes an image caption from this representation. Hence, our method conducts \textit{image-to-graph-to-text} generation by approaching the subtasks of \textit{image-to-graph} and \textit{graph-to-text} in an isolated fashion. To achieve this, we use two neural network architectures that focus on each task independently, and stack both architectures together once they have been trained.

\subsection{Scene graph generation}
We initially aim to solve the problem of \textit{image-to-graph} generation, i.e., generating a scene graph consisting of objects and visual relationships present in a given input image. Formally, our scene graph generator crafts a scene graph $G_I = (V,E)$ for an input image $I$ that consists of a set of nodes $V$ and corresponding directed edges $E \subseteq V \times V$. Each node $v$ is associated with a label $\kappa(v)$, representing an object in an image (e.g., \textit{car, person, building}). Likewise, each edge $e = (v_i, v_j) \in E$ is assigned a label $\kappa(e)$ denoting a relationship between the two objects $\kappa(v_i)$ and $\kappa(v_j)$ (e.g., \textit{above, on}).

In order to generate a graph $G_I$ from raw input pixels $I$, we make use of an existing scene graph generation model called \textsc{MotifNet}~\cite{zellers2018scenegraphs}. This method represents a scene graph as a triplet $G_M = (B, O, R)$, with $B = \{b_1, \dots, b_n\}, b_i \in \mathbb{R}^4$ a set of bounding boxes, $O = \{o_1, \dots, o_n\}$ a set of objects where each $o_i$ corresponds to a bounding box $b_i$, and $R = \{r_1, \dots, r_m\}$ a set of relationships where each relationship $r_k$ is a triplet $r_k = ((b_i, o_i), (b_j, o_j), x_{i\to j})$. Here, $(b_i, o_i), (b_j, o_j) \in B \times O$ represent the start and end node of the relationship and $x_{i\to j} \in \mathcal{R}$ denotes the relationship between both nodes from all possible relationships $\mathcal{R}$. Based on this scene graph representation, \textsc{MotifNet} computes the probability $P(G_M \, |\, I)$ of observing graph $G_M$ given image $I$ by decomposing it into three parts:
\begin{displaymath}
P(G_M \,|\,I) = P(B \,|\, I) \cdot P(O \,|\, B,I) \cdot P(R\,|\,B,O,I)
\end{displaymath}

\citet{zellers2018scenegraphs} model $P(B \,|\, I)$ with the \textit{Faster R-CNN}~\cite{ren2015} bounding box detection model. They then employ two LSTM networks~\cite{hochreiter97} to estimate the bounding box labels $P(O \,|\, B,I)$. Subsequently, the authors employ a bidirectional LSTM to compute the relationships between objects identified by the object detector as denoted by $P(R\,|\,B,O,I)$. To do so, all possible pairs of detected objects are taken into account and the LSTM computes a probability distribution over all potential relationships in $\mathcal{R}$ for each pair of objects.

\subsection{Graph-to-text generation}
Once we have generated a graph representation $G_I = (V,E)$ for an input image $I$, we utilize an LSTM decoder with an additional attention mechanism over the graph to generate an output sequence in natural language. Our architecture receives a set of graph nodes $V$ and maps each node $v \in V$ to an embedding representation $\mathbf{v} \in\mathbb{R}^D$ corresponding to its node label $\kappa(v)$. Hence, in order to represent visual relationships in this setup, we first transform our graph $G_I$ to a new representation $G_I^\prime = (V^\prime, E^\prime)$ that differs from $G_I$ in that each edge label is now assigned an individual node in the graph, i.e., for each $e=(v_i, v_j) \in E$ we create a new node $v^\prime$ such that $\kappa(e) = \kappa(v^\prime)$ and add edges $e_i^\prime = (v_i, v^\prime), e_j^\prime = (v^\prime, v_j)$ to $E^\prime$ with $\kappa(e_i^\prime) = \kappa(e_j^\prime) = \texttt{None}$.

Our method then applies an LSTM to the matrix $\mathbf{V} = [\mathbf{v}_1, \dots, \mathbf{v}_n] \in \mathbb{R}^{D \times n}$ of each node's embedding representation. To do so, we follow~\citet{xu16} and first initialize the LSTM's hidden and cell states as
$$
\mathbf{h}_0 = \psi_h\left(\frac{1}{n} \sum_{i=1}^n \mathbf{v}_i\right) \quad \mathbf{c}_0 = \psi_c\left(\frac{1}{n} \sum_{i=1}^n \mathbf{v}_i\right),
$$
where $\psi_h$ and $\psi_c$ are two independent multilayer perceptrons. Based on this initial conditioning, we then decode the image caption by sampling from
$$
p(\mathbf{y}_t\,|\,\mathbf{y}_{t-1}, \mathbf{V}) \propto \exp(\mathbf{P}_o\tanh(\mathbf{E}_W \mathbf{y}_{t-1} + \mathbf{P}_h\mathbf{h}_t))
$$
at each time step $t$, thereby also following~\citet{xu16}. Here, $\mathbf{E}_W \in \mathbb{R}^{D \times V}$ represents our word embedding matrix ($V$ is the vocabulary size), $\mathbf{y}_{t-1} \in \{0,1\}^V$ is a one-hot representation of the model's prediction at time step $t-1$ (or a special start token at $t=0$), $\mathbf{h}_t$ is the LSTM's hidden state at time step $t$ and $\mathbf{P}_o, \mathbf{P}_h$ are trainable parameter matrices. In the remainder of this work, we refer to the combination of our graph encoder and this type of decoder as \glstm{}.

Our second model variant incorporates an additional attention mechanism operating over the latent graph representation $\mathbf{V}$ at each time step $t$ of the LSTM. We adapt~\citet{xu16}'s approach for image captioning with visual attention and replace the latent image representation with our graph nodes, thus enabling our model to refer back to the graph representation and identify the most salient nodes at each time step during the generation of the output sequence. We call this extended approach $\text{\textbf{G-LSTM}}_{\mathcal{A}}$.

\subsection{Encoding visual relationships}
The aforementioned \glstmatt{} does not explicitly incorporate the visual relationships between objects as represented in the scene graph, but instead only processes all object and relationship nodes to generate an image caption. We thus experiment with the incorporation of an additional graph encoder that maps the initial graph representation $\mathbf{V} = \{\mathbf{v}_1, \dots, \mathbf{v}_n\}$ to an output representation $\mathbf{V}^\prime = \{\mathbf{v}_1^\prime, \dots, \mathbf{v}_n^\prime\}$. The task of this graph encoder is to encode relational information for each graph node into its corresponding graph embedding to provide the decoder with semantic dependencies between individual nodes in the graph. Additionally, the encoder has the ability to process indirect connections between entities in order to contextualize global relationships between entities that are indirectly connected through multiple edges. \textit{Graph Attention Networks} (GAT;~\citet{velickovic18}) represent a gradient-based approach that transforms an input graph by individually attending over each node's neighborhood to encode relational information into the resulting node representations. For a given input graph $G=(V,E)$, we then define a graph representation $\mathbf{G} = (\mathbf{V}, \mathbf{E})$, where
$$
\mathbf{E} = \{\{\mathbf{v}_i, \mathbf{v}_j\} \, | \, (v_i, v_j) \in E \lor (v_j, v_i) \in E\}
$$
represents the set of undirected edges in $G$ corresponding to $E$.

GAT layers transform the node representations by computing attention over their neighborhoods. Formally, let $\mathcal{N}_i$ denote the neighborhood of a node embedding $\mathbf{v}_i \in \mathbf{V}$. A GAT layer $\phi: \mathbb{R}^D \to \mathbb{R}^{D^\prime}$ then transforms each $\mathbf{v}_i$ to $\mathbf{v}_i^\prime$ by computing
$$
\mathbf{v}_i^\prime = \phi(\mathbf{v}_i) = \sigma \left(
\sum_{\mathbf{v}_j \in \mathcal{N}_i} \alpha_{ij} \mathbf{W}\mathbf{v}_j\right).
$$
Here, $\sigma$ represents the sigmoid function and $\alpha_{ij}$ is an attention coefficient with respect to the nodes $\mathbf{v}_i$ and $\mathbf{v}_j$. We follow~\citet{velickovic18} and set
$$
\alpha_{ij} = \frac{
\exp(\mathrm{LR}(\mathbf{a}^T[\mathbf{W}\mathbf{v}_i || \mathbf{W}\mathbf{v}_j]))
}{
\underset{\mathbf{v}_k\,\in\,\mathcal{N}_i}{\sum} \exp(\mathrm{LR}(\mathbf{a}^T[\mathbf{W}\mathbf{v}_i || \mathbf{W}\mathbf{v}_k]))}
$$
where $\mathbf{W} \in \mathbb{R}^{D^\prime \times D}, \mathbf{a} \in \mathbb{R}^{2D^\prime}$ are trainable weight matrices, $||$ represents vector concatenation and LR$(\cdot)$ denotes the LeakyReLU activation function. In our experiments, we define $\mathcal{N}_i := \{\mathbf{v} \in \mathbf{V} \,|\,\{\mathbf{v}_i,\mathbf{v}\} \in \mathbf{E}\} \cup \{\mathbf{v}_i\}$ to ensure a direct connection between an input node $\mathbf{v}_i$ and its transformation $\phi(\mathbf{v}_i)$ in each GAT layer.

Our final graph encoder then consists of multiple GAT layers that are executed sequentially to transform the node embedding representations with respect to their relationships in the graph.
Once our encoder has processed the initial graph embedding representation, we then feed our \glstm{} models with this representation and train the entire model in an end-to-end fashion. We denote both model variants with \glstmenc{} and \glstmencatt{}.

\subsection{Conventional image captioning baselines}
To provide a comparison between our approach and the conventional CNN-LSTM image captioning, we adapt~\citet{xu16}'s method. We preprocess each input image using the VGG19 network~\cite{simonyan15} pre-trained on ImageNet, and condition our LSTM language model on the $14 \times 14 \times 512$ feature representation emitted by the fifth layer of VGG19 before applying max-pooling. Analogously to the graph-to-text models, we furthermore experiment with an additional visual attention mechanism operating over the input image (see~\citet{xu16}). We denote both approaches with \pixeltocap{} and \pixeltocapatt{}.

Figure~\ref{fig:architecture} provides an overview and comparison of both the \glstmatt{} and \pixeltocapatt{} models. Both follow a similar technique of firstly encoding an input image by transforming it to a latent representation. This latent representation is then used to decode the corresponding image caption using an attention mechanism. However, a major difference between \pixeltocapatt{} and \glstmatt{} is that the latent representation of the latter (i.e., the scene graph) allows humans to explicitly observe which visual and contextual information have been extracted from the image. This property is not given for the \pixeltocapatt{} approach, since the latent representation emitted by the CNN is highly abstracted and hence less interpretable.

\section{Experiments}
We conduct a series of experiments on a subset of the Visual Genome~\cite{krishna17} and MS COCO~\cite{mscoco2014} datasets consisting of images accompanied by bounding boxes, scene graphs and individual image captions.

We use the BLEU~\cite{papineni02} and METEOR~\cite{denkowski14} evaluation metrics to measure the performance of our proposed approaches and to be able to compare them to existing methods for image caption generation. Both metrics have been used in a variety of studies related to image caption generation~\cite[e.g.,][]{xu16,vinyals15,lu2017}.

\begin{figure}[!ht]
        \begin{minipage}{.48\columnwidth}
            \includegraphics[width=\columnwidth]{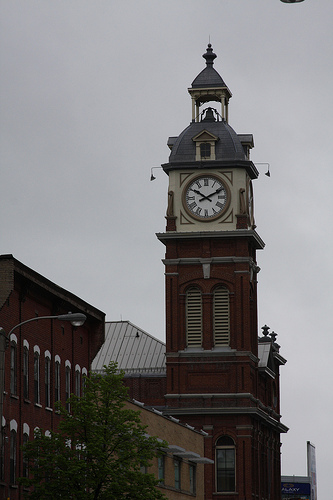}
        \end{minipage} \hfill
        \begin{minipage}{.48\columnwidth}
            \includegraphics[width=0.75\columnwidth]{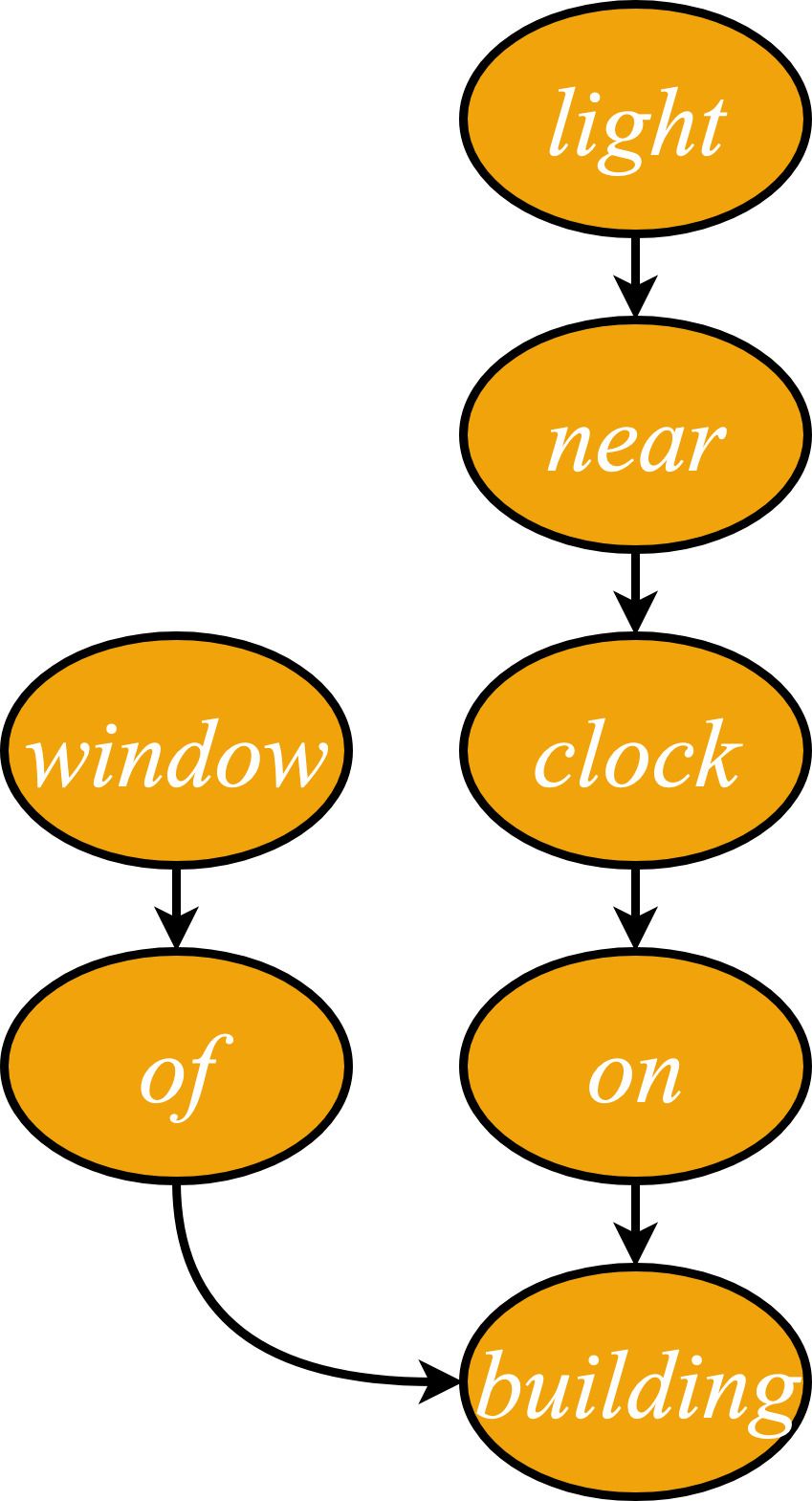}
        \end{minipage} \hfill

        \begin{enumerate}\itemsep -3ex
        \item\textit{a clock tower is in the gray sky.}\\
        \item\textit{clock tower ascending into overcast sky from buildings below.}\\
        \item\textit{a church tower that has a clock for the public.}\\
        \item\textit{brick building with clock tower in urban setting.}\\
        \item\textit{a tall clock tower near a building.}
        \end{enumerate}

        \caption{An example triplet of our generated dataset. The image is present in both the Visual Genome and MS COCO datasets. The scene graph is taken from a modified set of scene graphs from Visual Genome~\cite{Xu2017SceneGG} and the five image captions are taken from MS COCO.}
        \label{fig:example-dataset}
    \end{figure}

\subsection{Datasets}
Our dataset consists of a subset of all 51,498 images at the intersection of the Visual Genome and MS COCO datasets. First, we split the 51,498 images into a test set of 5,000 images, a validation set of 1,000 images and a training set of 45,498 samples. Operating on the intersection of VG and MS COCO allows us to craft triplet samples consisting of an image, a corresponding scene graph and a list of captions describing the image. In order to be as consistent as possible with the existing literature on scene graph generation, we then match all dataset samples with a modified Visual Genome dataset as explained in~\citet{Xu2017SceneGG}, considering only the 150 most common object categories and 50 most common relationships.
As each image in the training set is on average accompanied by 5.002 captions sourced from MS COCO,
the graph-to-text generation module can be trained with a total amount of 221,792 (\textit{scene graph, caption}) pairs. 

An example for a single element from our generated dataset (image, scene graph and captions) can be found in Figure~\ref{fig:example-dataset}.

During validation and testing, we evaluated our model's predictions using all available captions for a given image.

\begin{table}[]
\centering
\resizebox{\columnwidth}{!}{%
\begin{tabular}{lccccc}
\toprule
\textbf{Model} & \textbf{B-1} & \textbf{B-2} & \textbf{B-3} & \textbf{B-4} & \textbf{METEOR} \\ \midrule
$\text{\pixeltocap{}}$ & 65.58 & 43.93 & 29.58 & 20.40 & 22.90 \\
\pixeltocapatt{} & 66.09 & 44.04 & 29.32 & 19.96 & 22.65 \\
\midrule
\glstm{} & 67.29 & 45.47 & 30.48 & \textbf{20.85} & 23.79 \\
\glstmatt{} & \textbf{67.71} & \textbf{45.95} & \textbf{30.63} & 20.70 & \textbf{23.87} \\
\glstmenc{} & 66.30 & 43.56 & 28.33 & 18.82 & 22.75 \\
\glstmencatt{} & 65.63 & 43.69 & 28.81 & 19.48 & 23.33 \\ \bottomrule
\end{tabular}%
}
\caption{BLEU and METEOR scores for all trained models when evaluated on the test set. The \pixeltocap{} and \pixeltocapatt{} models were evaluated on the VGG image representations, and all \glstm{} models were evaluated on the scene graphs generated by \textsc{MotifNet} (trained and tuned on our training and validation sets, respectively). Bold values indicate best performances for each evaluation criterion across all models.}
\label{tab:all-gen-results}
\end{table}

\subsection{Implementation details and training}
We trained the individual submodules responsible for the image-to-graph and graph-to-text generation independently on the aforementioned datasets.

The scene graph generator was trained by strictly following~\citet{zellers2018scenegraphs}'s approach to train their proposed model.\footnote{We followed the authors' instructions on \url{https://github.com/rowanz/neural-motifs}.} This approach consists of three phases. First, a Faster R-CNN object detector with a VGG backbone is pre-trained in isolation to learn the extraction of objects and corresponding bounding boxes from images. We adhered to the architecture and parameter setup as explained in their work, and trained the detector for 50 epochs. After training the object detector, we trained the \textsc{MotifNet} module for 26 epochs without modifying the authors' implementation setup (this includes the adaptation to scene graph detection as explained in~\citet{zellers2018scenegraphs}, Section 5.2). For the graph-to-text models, we tokenized all sequences used during training using the NLTK \texttt{tokenize} package~\cite{bird04}. We did not exclude infrequent vocabulary tokens during our analysis. All reported models were trained using the \textit{Adam} optimizer~\cite{kingma2014} with a learning rate of $1 \cdot 10^{-4}$. In terms of model regularization, we used  \textit{dropout}~\cite{srivastava14} in both the encoder and the decoder during training. In the encoder, we added a dropout mechanism with a rate of 0.25 at each GAT layer directly before computing the weighted sum of the transformer graph inputs. In the decoder, we adhered to the use of dropout as realized by~\citet{xu16} and used a dropout rate of 0.5. Moreover, we use \textit{batch normalization}~\cite{ioffe15} in the LSTM decoder by normalizing the encoder outputs before transforming them to the LSTM's initial hidden and cell states. Our graph encoder consists of two consecutive GAT layers that are operating on a dimension of $D=512$. We set the dimension of the trainable graph and word embeddings to the same size and utilize a single-layer LSTM with 1024 hidden units as decoder.

We trained our two conventional image captioning baselines \pixeltocap{} and \pixeltocapatt{} with the same hyperparameter settings.

\subsection{Tuning \textsc{MotifNet} on the validation set}
For a given input image, the trained \textsc{MotifNet} generates both a list of detected bounding boxes along with their predicted labels as well as a list of relationship predictions between the identified objects. In detail, it outputs a probability distribution over all possible 50 relationship predicates for each pair of predicted objects. 
However, the \textit{Faster R-CNN} object detector predicts certain bounding box labels with low confidence values which might result in scene graph representations with high model uncertainty. To account for this problem, and to limit the size (i.e., number of nodes) of the generated scene graphs, we experimented with various confidence threshold values representing lower bounds for the confidence values of the object detector to be considered a valid object of an image. Specifically, we considered the confidence thresholds 0.2, 0.4, 0.6, and 0.8 for our trained models. For each of the four \glstm{} model variants, we thus evaluated to what extent these different confidence thresholds affected the overall model performances (in terms of METEOR) by experimenting how the model variants perform on the validation set with each of the parameter values.
Our results suggest that the \glstm{},
\glstmencatt{} and \glstmenc{} models exhibit their best performance with a confidence threshold of 0.4, while the \glstmatt{} variant performs best with a confidence threshold of 0.2.

Once we have identified all valid predicted objects present in the image, we selected the graph's relationships by considering all relationships between valid objects suggested by \textsc{MotifNet} and assigned the predicate with highest probability as the relationship label.

Finally, we removed all duplicate nodes and identical relationships from the crafted scene graph. If a generated scene graph exceeds the maximum size (i.e., number of nodes) of the graphs used during training, we limit the graph's size to this maximum size by removing the object nodes exhibiting the lowest prediction confidences. Moreover, if a scene graph consists of less than two predicted object nodes, we ignore the sample during testing.

\begin{table}[]
\centering
\resizebox{\columnwidth}{!}{%
\begin{tabular}{lccccc}
\toprule
\textbf{Model} & \textbf{B-1} & \textbf{B-2} & \textbf{B-3} & \textbf{B-4} & \textbf{METEOR} \\ \midrule
\glstm{} & 69.09 & 47.85 & 32.76 & 22.77 & 24.80 \\
\glstmatt{} & \textbf{69.48} & \textbf{48.31} & \textbf{33.16} & \textbf{22.91} & \textbf{24.81} \\
\glstmenc{} & 67.65 & 45.74 & 30.51 & 20.77 & 23.79 \\
\glstmencatt{} & 68.06 & 46.87 & 31.79 & 21.90 & 24.56 \\ \bottomrule
\end{tabular}%
}
\caption{BLEU and METEOR scores for all graph-based models when evaluated on the ground-truth scene graphs in the test set. Bold values indicate best performances for each evaluation criterion across all models.}
\label{tab:all-gt-results}
\end{table}

\begin{figure}[!ht]
        \begin{minipage}{.48\columnwidth}
            \vskip 0.2cm
            \begin{center}\textbf{Image}\end{center}
            \includegraphics[width=\columnwidth]{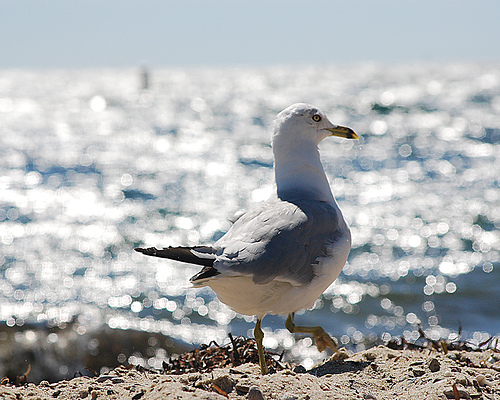}
        \end{minipage} \hfill
        \begin{minipage}{.48\columnwidth}
            \vskip -0.4cm
            \begin{center}\textbf{Generated graph}\end{center}
            \vskip 0.3cm
            \includegraphics[width=\columnwidth]{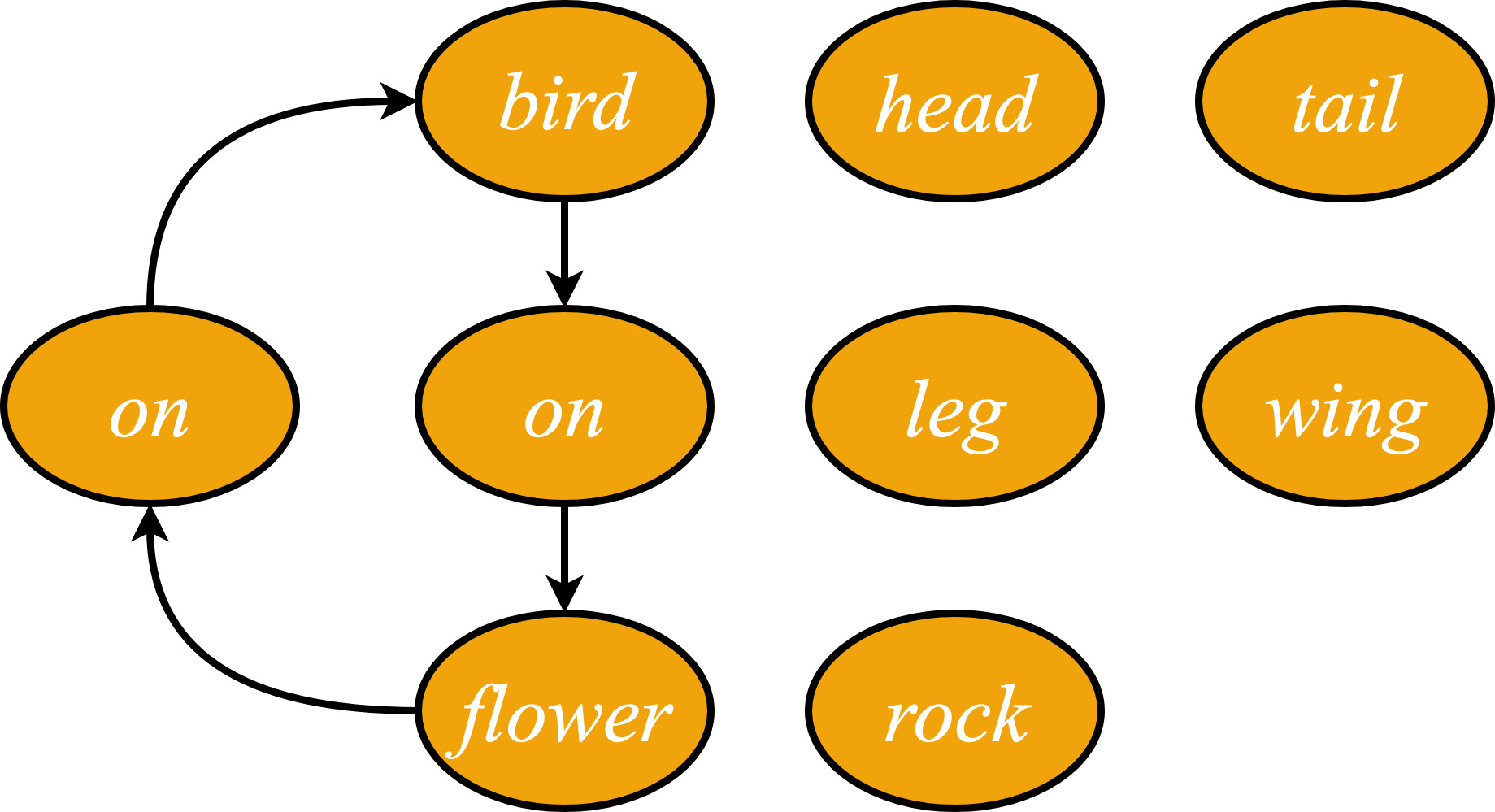}
        \end{minipage} \hfill
        \vskip 0.3cm
        \glstmatt{}: \textit{a bird is perched on a rock in the water.} \\
        \pixeltocapatt{}: \textit{a bird is flying in the air on a beach.}
        \caption{Comparison of generated image captions from both the \glstmatt{} and the \pixeltocapatt{} models. The graph shown on the right-hand side has been generated from the input image (left-hand side) using the trained \textsc{MotifNet}. The sequences below the image and graph represent the predicted captions for both systems.}
        \label{fig:example-result}
    \end{figure}

\subsection{Results}
Quantitative results of all our models can be found in Table~\ref{tab:all-gen-results}. The results of all graph-based models are based on the scene graphs generated by \textsc{MotifNet}, which we trained before on our new dataset. The results in Table~\ref{tab:all-gen-results} show that both the \glstmatt{} and the \glstm{} outperform both the \pixeltocap{} and \pixeltocapatt{} in every metric, indicating that our proposed models represent a suitable alternative to the conventional image captioning approaches. Figure~\ref{fig:example-result} shows qualitative results of the \pixeltocapatt{} and \glstmatt{} approaches in comparison, showing that our model is able to produce accurate captions even in the presence of imperfect auto-generated scene graphs. Furthermore, it is interesting to observe that the additional graph encoder operating over the input scene graph leads to performance decreases of our \glstm{} model. In addition to that, for both the conventional and the captioning model based on scene graphs, the attention mechanism operating on the decoding LSTM only slightly improves the overall model performance across our evaluation metrics.

To further assess the performance of our models when operating on generated scene graphs, we provide metrics for all model variants when evaluated on the ground-truth gold scene graphs as provided in the Visual Genome dataset in Table~\ref{tab:all-gt-results}. Our models exhibit even higher performances when evaluated on the gold scene graphs, indicating that our method has the potential to benefit from future progress in the field of scene graph generation.

\subsection{Limitations}
The presented method imposes a number of limitations that we would like to address in the following paragraph. First, our image-to-graph-to-text model utilizes a scene graph generation model that is restricted to predicting only 150 different object labels and 50 different edge labels. This represents a notable limitation to the model since it is explicitly trained to predict diverse English sentences from only a small subset of semantic concepts. Nevertheless, the fact that our proposed methods outperform conventional image captioning approaches (which do not have this additional constraint) suggests that the model still learns to predict semantic concepts outside of the 200 given ones in context, and achieves to reasonably generate other concepts that are likely to occur in the context of certain objects and relationships as represented by the scene graph. 

Moreover, it is worth mentioning that our proposed approach arguably requires a larger amount of computational resources to be trained properly when compared to conventional image captioning methods. In addition to that, the current study does not investigate the potential of our proposed architecture when trained in an end-to-end fashion, i.e., by developing a single pipeline that processes an input image, generates a scene graph representation and then uses this representation to create a corresponding image caption. At this point we would like to encourage other researchers focusing on image captioning to further explore the potential of explicitly incorporating visual objects and relationships with respect to this problem.

\section{Conclusion}
In this work, we proposed a supervised learning approach to generate image captions by explicitly leveraging detected objects and visual relationships. Our suggested model consists of a simple two-step procedure that first generates a scene graph representation from a given image and then uses this representation to generate an image description in natural language. Empirical results on a newly-generated dataset consisting of samples from the intersection of Visual Genome and MS COCO demonstrate the superiority of our model when compared to conventional image captioning approaches, indicating that our method provides a fruitful ground to further advance the task of image captioning.

\section*{Acknowledgements}
We gratefully acknowledge a Ph.D. scholarship awarded to the second author by the German Academic Scholarship Foundation (Studienstiftung des deutschen Volkes). This work was supported by the BMBF as part of the project MLWin (Grant No. 01IS18050) as well as the Munich Center for Machine Learning (Grant No. 01IS18036B).

\bibliography{acl2021}
\bibliographystyle{acl_natbib}

\end{document}